\documentclass[conference]{IEEEtran}

\IEEEoverridecommandlockouts                              

\usepackage{times}

\usepackage{multicol}
\usepackage[bookmarks=true]{hyperref}

\usepackage{tikz}
\usepackage{comment}
\usepackage{amsmath,amssymb} 
\usepackage{color}
\usepackage{wrapfig}

\usepackage{url}
\usepackage{graphicx}
\usepackage{booktabs}
\usepackage{multirow}
\usepackage{vcell}

\usepackage{array}
\usepackage{makecell}
\usepackage{rotating}
\usepackage{arydshln}
\usepackage{soul} 
\usepackage{tabularray}

\title{\LARGE \bf
SlotGNN: Unsupervised Discovery of Multi-Object \\Representations and Visual Dynamics}

\author{Alireza Rezazadeh$^{1}$, Athreyi Badithela$^{2}$, Karthik Desingh$^{2*}$, Changhyun Choi$^{1*}$
\thanks{$^{1}$ Department of Electrical and Computer Engineering, University of Minnesota, Minneapolis, MN 55414 USA ({\tt\small rezaz003@umn.edu}, {\tt\small cchoi@umn.edu}).}%
\thanks{$^{2}$ Department of Computer Science and Engineering, University of Minnesota, Minneapolis, MN 55414 USA ({\tt\small kdesingh@umn.edu}, {\tt\small badit004@umn.edu}).}
\thanks{$^{*}$ Equal contributions.}
}

\usepackage{etoolbox}
\makeatletter
\patchcmd{\@makecaption}
  {\scshape}
  {}
  {}
  {}
\makeatother

\begin{document}
\maketitle

\begin{abstract}
Learning multi-object dynamics from visual data using unsupervised techniques is challenging due to the need for robust, object representations that can be learned through robot interactions. This paper presents a novel framework with two new architectures: SlotTransport for discovering object representations from RGB images and SlotGNN for predicting their collective dynamics from RGB images and robot interactions. Our SlotTransport architecture is based on slot attention for unsupervised object discovery and uses a feature transport mechanism to maintain temporal alignment in object-centric representations. This enables the discovery of slots that consistently reflect the composition of multi-object scenes. These slots robustly bind to distinct objects, even under heavy occlusion or absence. Our SlotGNN, a novel unsupervised graph-based dynamics model, predicts the future state of multi-object scenes. SlotGNN learns a graph representation of the scene using the discovered slots from SlotTransport and performs relational and spatial reasoning to predict the future appearance of each slot conditioned on robot actions. We demonstrate the effectiveness of SlotTransport in learning object-centric features that accurately encode both visual and positional information. Further, we highlight the accuracy of SlotGNN in downstream robotic tasks, including challenging multi-object rearrangement and long-horizon prediction. Finally, our unsupervised approach proves effective in the real world. With only minimal additional data, our framework robustly predicts slots and their corresponding dynamics in real-world control tasks.
Our project page: \href{http://bit.ly/slotgnn}{bit.ly/slotgnn}.

\end{abstract}

\IEEEpeerreviewmaketitle

\section{Introduction}

Studies suggest that the human visual system identifies conceptually distinct visual features, indexes their locations \cite{pylyshyn1989role}, and utilizes this information as the foundation for higher-level cognitive processes, such as comprehending and interacting effectively with the world \cite{marr2010vision}.
A similar principle guides many robotic systems for goal-directed motor planning. 
In multi-object manipulation, early approaches aimed to directly project the image observation into a unified lower-dimensional space to infer the dynamics \cite{finn2017deep, agrawal2016learning}. However, such strategies do not reflect the inherent structure of a multi-object system and lack object-level predictions. This limitation not only impedes the model's ability to learn object interactions but also results in inaccurate dynamics predictions.
Addressing this limitation, recent methods build dynamics models by decomposing the observation into object-specific lower-dimensional latents and subsequently learning dynamics within these ``object-centric'' representations \cite{driess2023learning, ye2020object, qi2020learning, minderer2019unsupervised, qi2020learning}. For multi-object systems, recent studies emphasize the effectiveness of learning object-centric representations to enhance the accuracy and sample efficiency of dynamic models \cite{driess2023learning, ye2020object}. This category of models follows a natural formulation by first learning to represent a scene as a set of object-centric features and then learning the dynamics among them. 

In robotics, unsupervised learning of object dynamics is a key challenge particularly given its significance in model-based action planning for real-world applications. Nevertheless, the majority of existing methods of learning multi-object dynamics heavily rely on ground-truth information, including object pose \cite{battaglia2016interaction, sanchez2018graph, ye2020object} and segmentation masks \cite{ye2020object, driess2023learning}. This substantially restricts the applicability of such solutions in real-world settings where comprehensive ground-truth information is often unavailable. To address this challenge, our work focuses on discovering unsupervised object representations in multi-object scenarios and harnessing these representations to understand their dynamics. Our primary contributions include:

\begin{figure}[t]
    \vspace{4mm}
    \centering
    \includegraphics[width=1\linewidth]{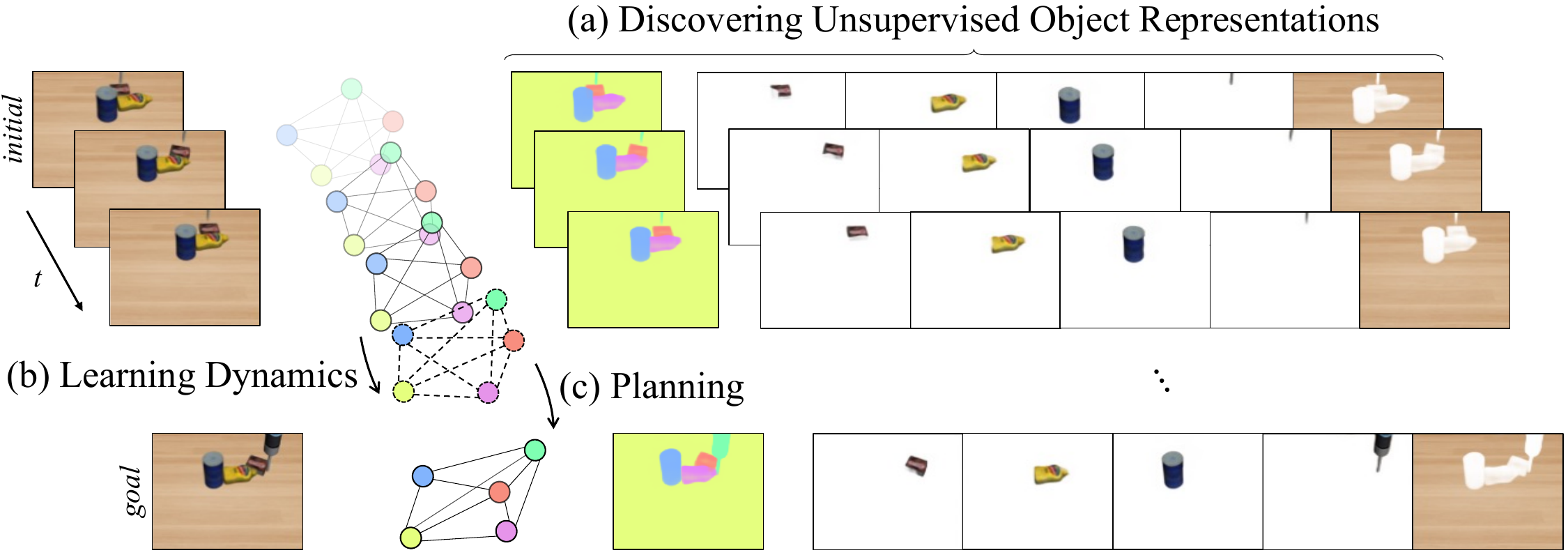}
    \caption{
    Overview of our unsupervised framework. (a) We introduce \textbf{SlotTransport} to identify temporally-aligned, object-centric slots, that each consistently represents a unique visual element. (b) We introduce \textbf{SlotGNN}, a graph-based model that learns scene dynamics from slots and predicts future states based on the robot's action. (c) Our unsupervised approach facilitates planning to transition from an initial state to a goal image without requiring extensive ground-truth supervision.
        } 
    \vspace{-5mm}
    \label{fig:banner}
\end{figure}

\begin{figure*}[ht]
    \centering
    \includegraphics[width=0.9\linewidth]{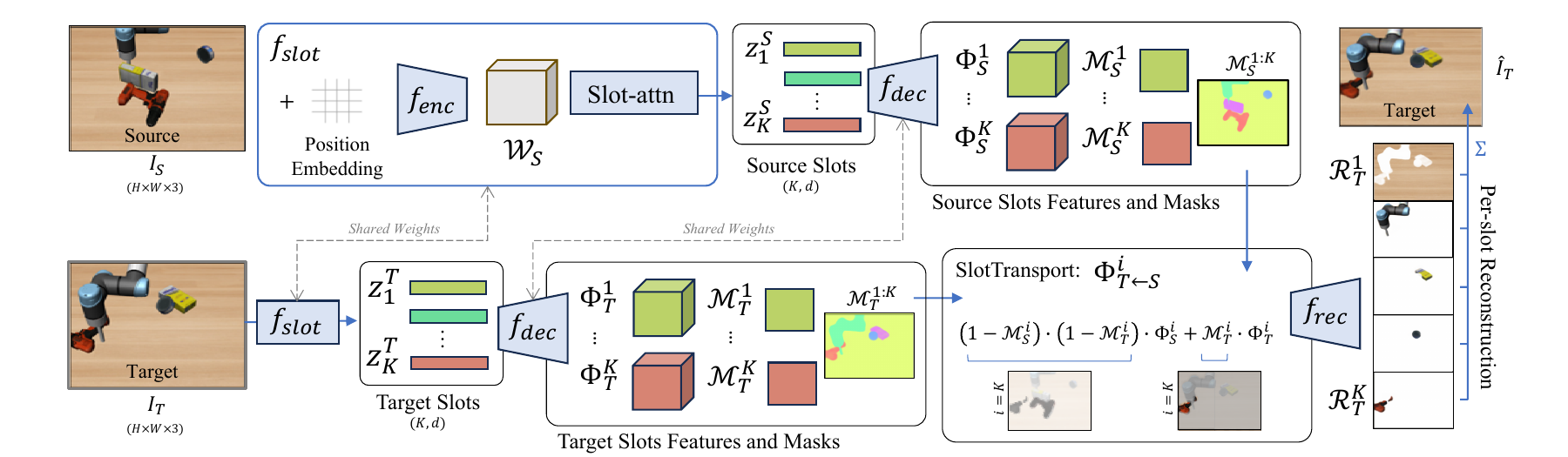}
\caption{
\textbf{SlotTransport}: Unsupervised Multi-Object Discovery. 
From an RGB image, $f_{slot}$ identifies object-centric slots $z_{1:K}$. Through slot attention \cite{locatello2020object}, slots bind to visual features, and $f_{dec}$ produces feature maps $\Phi^i$ and masks $\mathcal{M}^i$. Temporal alignment is ensured by transporting slot features between source and target images. $f_{rec}$ reconstructs each object slot $\mathcal{R}^i_T$, which together compose the target image. The model is trained using only reconstruction error. During inference, objects are reconstructed from a single image using learned slots.
}
\label{fig:arch-slotgtrans}
\end{figure*}


\noindent(1) We introduce \textbf{SlotTransport} for unsupervised object discovery, a novel architecture that refines object-centric representation learning through slot attention \cite{locatello2020object}. Utilizing a feature transport mechanism, SlotTransport ensures temporal alignment of the object-centric representations. The discovered slots capture scene composition, each depicting a visual entity in a multi-object scene, such as objects, the background, and the robot. Notably, each slot maintains a consistent association with a distinct object, even when it's occluded or absent.

\noindent(2) We propose \textbf{SlotGNN}, a novel unsupervised graph-based model for predicting multi-object scene dynamics from object-centric representations. SlotGNN uses slots identified by SlotTransport to synthesize the scene's future appearance based on the robot's actions. With the temporal alignment from SlotTransport, the scene transforms into a graph where each node consistently represents a slot, and edges capture the slot interactions. SlotGNN performs relational reasoning on the graph and learns to project the future appearance of each slot. 


\noindent(3) We examine the dynamics learned with SlotGNN in challenging downstream robotic tasks. We employ SlotGNN for challenging goal-directed multi-object rearrangement using pushing actions and long-horizon dynamics prediction. 

\noindent(4) Demonstrating the real-world applications of our unsupervised approach, we successfully transfer SlotTransport and SlotGNN, initially trained in simulation, to the real robot by collecting a minimal dataset of just 20 real robot demonstrations (5\% of the amount of simulated training data). 

\noindent Our results demonstrate the robustness of our unsupervised framework, particularly in downstream robotic applications and real-world scenarios. Our approach consistently predicts accurate multi-object representations and their corresponding dynamics. Throughout this paper, we will use the terms `slots' and `object-centric representations' interchangeably.

\section{Related Work}

\begin{figure*}[ht]
    \centering
    \includegraphics[width=1.00\linewidth]{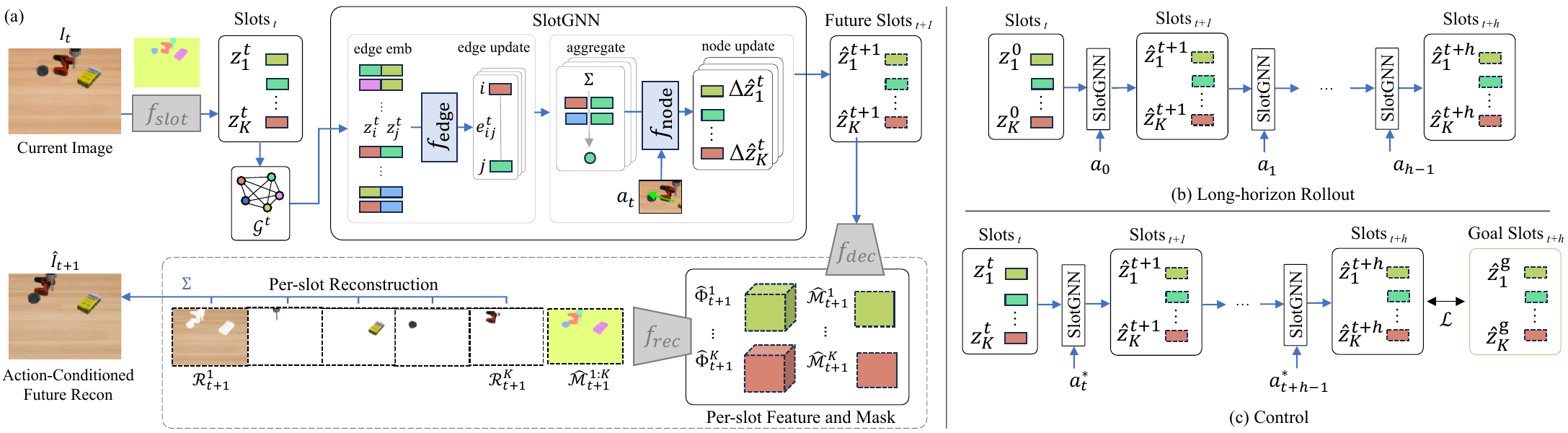}
\caption{
\textbf{SlotGNN}: Unsupervised Multi-Object Dynamics. 
(a) SlotGNN predicts slot changes \(\Delta z_i^t\) after applying a pushing action \(a_t\). Using SlotTransport's slots (\(f_{slot}\)), a graph $\mathcal{G}_t$ is formed with slots as nodes and slot interactions as edges. Edges and nodes are updated via \(f_{edge}\) and \(f_{node}\), resulting in next timestep slots \(z_i^{t+1}\). The next image \(\hat{I}_{t+1}\) is then reconstructed.
(b) With a sequence of robot actions, SlotGNN projects future multi-object dynamics and synthesizes future scenes. 
(c) SlotGNN also facilitates goal-directed planning to optimize actions towards a desired goal.
}
    \label{fig:arch-slotgnn}
\end{figure*}


\subsubsection*{Learning Multi-object Dynamics Model}

Early models for graph-based dynamics, such as Interaction Networks (IN) \cite{battaglia2016interaction, sanchez2018graph} and follow-up adaptations \cite{kipf2018neural, li2018propagation, sanchez2020learning}, represent a multi-object system with a graph where each node is an object ground-truth state (e.g., position, velocity, mass, friction). These models rely on explicit state information. However, for real-world robotic scenarios, obtaining ground-truth state data is infeasible.
Recent methods explored learning object representations. Each object is mapped to a lower-dimensional, object-centric representation. The representations are typically a combination of explicit ground-truth states, like position, bounding box, and mask, combined with implicit visual features \cite{yuan2022sornet, qi2020learning, watters2017visual, ye2020object, driess2023learning}. However, the primary assumption of the ground-truth state supervision limits their application in the real world. 
In contrast, our work introduces an unsupervised framework for learning multi-object scene dynamics based on discovering unsupervised slots. This eliminates the need for explicit ground-truth state supervision.

\subsubsection*{Unsupervised Object-centric Representation}
Our work builds on learning object-centric representations using slot attention \cite{locatello2020object}. Slot attention interfaces with visual outputs to generate a set of slots. For robotics applications, ensuring temporal consistency in these slots is vital for accurate scene dynamics understanding. This consistency is essential for formulating a planning objective or training loss. Thus, we explicitly incorporate a feature transport mechanism in our SlotTransport to maintain consistency across image pairs from different observation timesteps inspired by \cite{kulkarni2019unsupervised}. 
While slot attention has been recently adopted for object localization and behavior cloning \cite{heravi2023visuomotor}, the experiments were limited to basic untextured objects and did not consider learning dynamics that is required for online planning. 
On another front, while keypoint-based methods such as \cite{rezazadeh2023kinet} learned unsupervised multi-object dynamics, they face difficulties handling occlusions—a common challenge in robotics. In contrast, our SlotTransport reliably handles occlusions and consistently associates slots with specific objects.
\section{Methods}

\noindent Our framework has two main components:

(1) \textbf{SlotTransport}: An unsupervised multi-object discovery model that efficiently extracts robust and temporally consistent object-centric representation slots from multi-object scenes.

(2) \textbf{SlotGNN}: Building on top of the slot discovery, this unsupervised graph-based model learns the dynamics of the object-centric representations. Importantly, SlotGNN is conditioned on the robot's action which enables applications such as model-based action planning. 

\subsection{SlotTransport: Unsupervised Multi-Object Discovery}

The SlotTransport's role is to map the image to underlying object-centric representations. A detailed architecture of SlotTransport is shown in Fig. \ref{fig:arch-slotgtrans}. We build on the slot attention \cite{locatello2020object} to extract slots from image frames while ensuring temporal alignment of the slots. Given an RGB image $I$, a convolutional encoder $f_{enc}$ augmented with positional embeddings, maps the image to an intermediate representation of $\mathcal{W}\in\mathbb{R}^{h\times w\times c}$. Using the slot attention \cite{locatello2020object}, slots $z_{1:K}\in\mathbb{R}^{d}$ are derived that uniquely represent distinct portions of $\mathcal{W}$. 

We recognize that our ultimate goal of learning dynamics in an object-centric latent space requires temporal consistency of the slots. To explicitly enforce this, we introduce the transport mechanism in SlotTransport to establish temporal alignment in slots. Inspired by \cite{kulkarni2019unsupervised}, this mechanism transports slot features between a pair of source and target images $(I_S, I_T)$ sampled from a given scene. 

For each image, slots are extracted as $(z_{1:K}^S, z_{1:K}^T)$. First, using a single convolutional decoder $f_{dec}$, we decode each slot into a feature map $(\Phi^{1:K}_S, \Phi^{1:K}_T) \in\mathbb{R}^{h\times w\times m}$ and an alpha mask $(\mathcal{M}^{1:K}_S, \mathcal{M}^{1:K}_T) \in\mathbb{R}^{h\times w}$. These alpha masks serve as mixture weights to inpaint each slot's feature map from the target image onto the source image (see Fig. \ref{fig:arch-slotgtrans}). We produce a transported feature map $\Phi_{T\leftarrow S}$ by nullifying the source feature map outside the slot's predicted mask for both the target and source $(1-\mathcal{M}^i_T)\cdot(1-\mathcal{M}^i_S)\cdot\Phi^i_S$, followed by overlaying the masked target feature map $\mathcal{M}^i_T.\Phi^i_T$. Finally, a convolutional reconstruction module $f_{rec}$ reconstructs each slot as an RGB image $\mathcal{R}_T^i\in\mathbb{R}^{h\times w\times 3}$ based on the transported feature map. The reconstructed slots together reconstructed target image $\hat{I}_T$.

The transport mechanism in SlotTransport enforces temporal alignment between image pairs during training. Notably, the learned slots through SlotTransport consistently register to a unique object even under heavy occlusion or absence of an object. During inference, SlotTransport can discover and reconstruct slots from a single image by directly reconstructing the extracted per-slot features and masks. SlotTransport ensures that each slot's feature map aligns well with its mask for learning consistent object representation across time. Importantly, this temporal alignment is achieved without adding additional learnable parameters; the same $f_{slot}$ and $f_{dec}$ are used when processing both source and target images. 

\subsection{SlotGNN: Unsupervised Multi-Object Dynamics}
The main purpose of SlotGNN is to learn the dynamics and model interactions between the visual elements in a multi-object scene, such as the robot, objects, and the background. It does so using a graph-based representation, where each object-centric slot corresponds to a node in the graph.
Crucially, SlotGNN enables learning unsupervised multi-object dynamics, eliminating the need for supervised trajectory labels that require access to the system's ground-truth state. This feature becomes essential in real-world scenarios where obtaining accurate ground-truth data is challenging or impractical. Refer to the detailed architecture illustrated in Fig. \ref{fig:arch-slotgnn}-a.

Given an image \(I_t\) with its associated slots \(z_{1:K}^t\) discovered through SlotTransport, we construct a fully connected graph \(\mathcal{G}_t = (\mathcal{V}, \mathcal{E})\). Each node \(v_i \in \mathcal{V}\) in the graph represents a slot, and each edge \(e_{ij} \in \mathcal{E}\) represents the interaction between the pair of slots \(z_i, z_j\).
For each node, we associate an embedding \(n_i\) which is initialized with the slot representations \(z_i\). The edge embeddings, representing interactions, are initialized based on augmenting the connected slots representations. 
To process the information in the graph representation, SlotGNN employs a message-passing neural network architecture \cite{gilmer2017neural, sanchez2018graph} to update node and edge embeddings. Incoming information from neighboring nodes is aggregated to update each node's state, capturing the dynamics and interactions in the scene. 

The message-passing operation in the graph consists of two primary steps (see \ref{fig:arch-slotgnn}). First, the edge embeddings, $e_{ij}$, are updated based on their connecting node embeddings: 
\(e_{ij}' \leftarrow f_{edge}(e_{ij}, n_i, n_j)\).
Secondly, the node embeddings are updated using the updated edge embeddings associated with them and the robot action: \(n_k' \leftarrow f_{node}(n_k, \sum_{i \in \mathcal{N}(k)} e_{ik}', a_t)\).
Here, $f_{edge}$ and $f_{node}$ are multi-layer perception update functions for edges and nodes, respectively. $\mathcal{N}(k)$ denotes the neighbors of node $k$, which in the context of a fully connected graph is all other nodes.
To condition on external action, the robot action $a_t \in\mathbb{R}^4$, characterized as a point-to-point push vector in image coordinates, is integrated as an input to $f_{node}$ in SlotGNN. This ensures the learned dynamics are conditioned on the robot's action and can be used for planning in the downstream robotics control task.

After message-passing, the updated node embeddings are used to predict the evolution of slots \(\Delta \hat{z}_i^{t+1}\) in the next timestep conditioned on the action $a_t$. This allows for the rollout of the dynamics into the future and enables synthesizing the future appearance of the scene.

\begin{figure}[t]
    \centering
    \includegraphics[width=1.00\columnwidth]{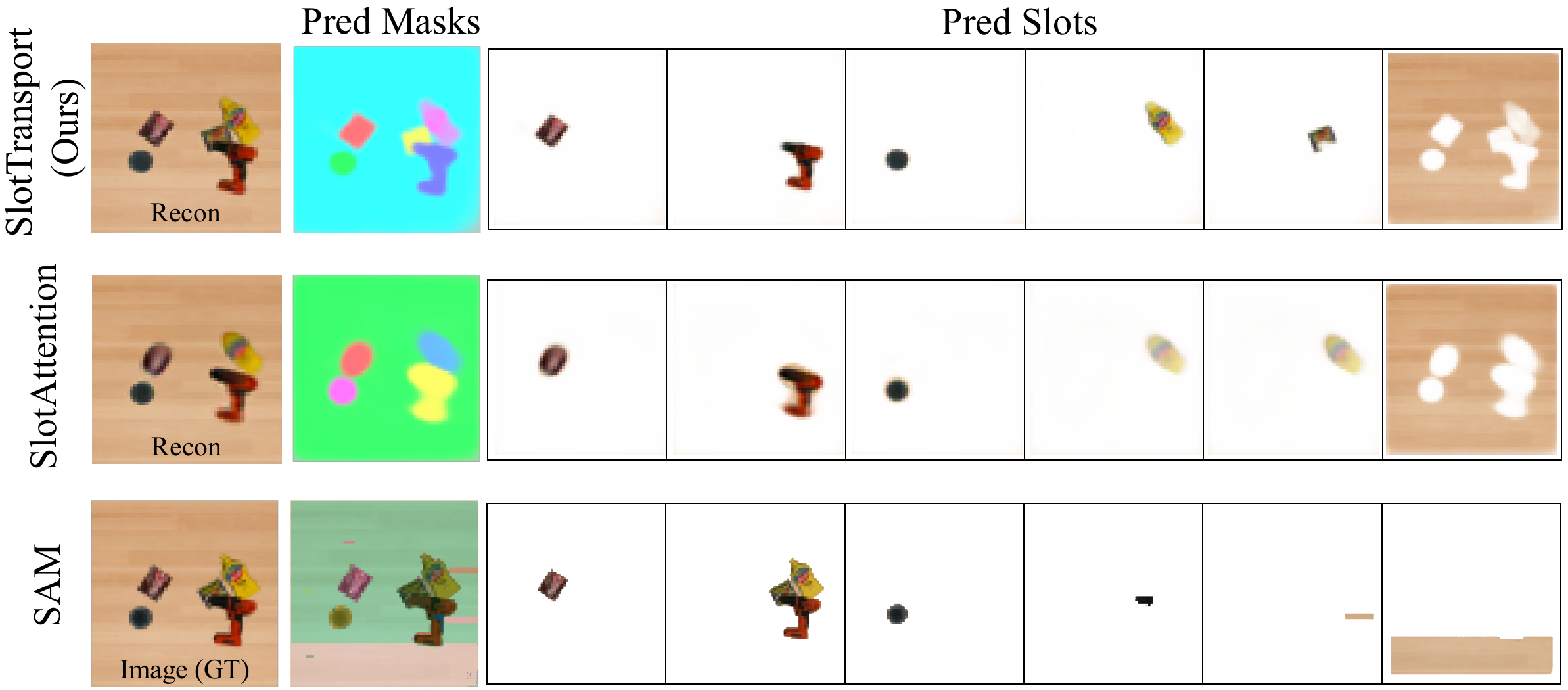}
    \caption{Visualizations of per-slot masks and reconstructions. SlotTransport exhibits superior performance in accuracy and consistency of object-centric representation, even under occlusion, compared to the SlotAttention baseline \cite{locatello2020object}. We also showcase predicted segments from SAM \cite{kirillov2023segment}.}
    \label{fig:segment-qual}
\end{figure}

\subsection{Training SlotTransport and SlotGNN}

SlotTransport is trained using only the image reconstruction error for supervision. The image reconstruction loss \(\mathcal{L}_{rec}(I_T, \hat{I}_T)\), is defined using a pixel-wise Mean Squared Error (MSE) between the target and reconstructed images. In Fig. \ref{fig:arch-slotgtrans}, modules with learnable parameters are distinctly highlighted in blue. Once SlotTransport is trained, it supervises the training of SlotGNN to learn visual dynamics. 

We use a slot prediction MSE loss to train SlotGNN, \(\mathcal{L}_{slot}({z}_{1:K}^{t+1}, \hat{z}_{1:K}^{t+1})\). This loss reduces the distance between the slots directly predicted from the next timestep image using SlotTransport ${z}_{1:K}^{t+1}$ and slots from the single-step dynamics with SlotGNN $\hat{z}_{1:K}^{t+1}$, as visualized in Fig. \ref{fig:arch-slotgnn}-a (modules with leanable parameters are highlighted in blue). Importantly, employing a per-slot prediction loss requires temporal alignment that is ensured through SlotTransport. Furthermore, we use the single-step slot dynamics to reconstruct the image and also minimize the image reconstruction MSE loss \(\mathcal{L}_{rec}(I_{t+1}, \hat{I}_{t+1})\).

\subsection{Long-Horizon Multi-Object Dynamics Rollout}
Given only an initial image frame \( I_0 \) and a sequence of robot pushing actions \( a_{0:h} \), SlotGNN can predict \(\hat{I}_{1:h}\) by recurrently running on the previous step's prediction.
As shown in Fig. \ref{fig:arch-slotgnn}-b, the single-step dynamics predictions are cascaded to predict slots over extended future horizons. This capability for accurate dynamics rollout is possible due to the temporal alignment achieved with SlotTransport. Furthermore, for any arbitrary future timestep, \( f_{rec} \) can synthesize an image of the scene from the predicted slots.

\subsection{Goal-Directed Planning}

Learning the scene dynamics facilitates goal-directed sequential action planning for multi-object manipulation. As shown in Fig. \ref{fig:arch-slotgnn}-c, with SlotGNN, we optimize robot actions to align a scene's state with a target goal image. This is pivotal when the robot interacts with several objects to reach a desired state. Given a scene image, we sample possible action sequences over a planning horizon \( h \geq 1 \) and forecast the slot representation \( \hat{z}^{t+h}_{1:K} \) by rolling the dynamics. Using Model-predictive control (MPC) \cite{camacho2013model}, the optimal action sequence is chosen by minimizing the slot loss \(\mathcal{L}_{slot}(z^T_{1:K}, z^G_{1:K})\), which quantifies the variance between predicted slots and the goal image slots.

\begin{figure}[t]
    \centering
    \includegraphics[width=\columnwidth]{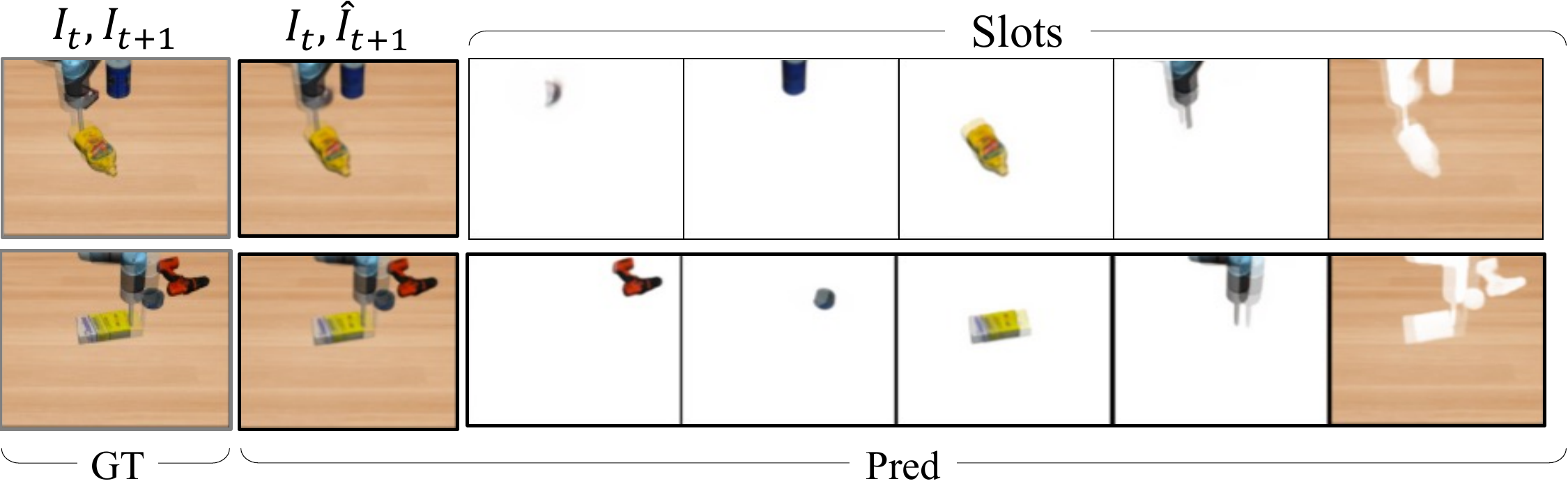}
    \caption{Examples of single-step dynamics prediction using SlotGNN. Given the current scene image and the robot's pushing action, our model precisely predicts the future state of each slot and synthesizes the future scene image.}
    \label{fig:qual-dyn}
\end{figure}

\section{Experiments}
We structure our experiments around: (1) How accurately and consistently do slots extracted by SlotTransport represent each visual element in the scene? (2) How effective is SlotGNN in predicting multi-object scene dynamics? (3) How well does our framework apply to downstream robotic tasks? 

\subsection{Data}
\textit{Simulation:} Using Mujoco \cite{todorov2012mujoco, robosuite2020}, we simulate a multi-object tabletop scene with YCB objects \cite{calli2015ycb} and a UR5e robot with a cylindrical end-effector. The robot performs planar pushing action, captured by an RGB camera. The data is formatted as image-action tuples \((I_t, a_t, I_{t+1})\) containing pre- and post-action images, and action vectors in the image coordinates. We generate $\sim 750$ episodes $\times 20$ steps of random pushes for a given subset of objects. SlotTransport is trained by randomly sampling target and source images across all episodes. We then use the learned SlotTransport to discover slots and train SlotGNN on the image-action tuples. Evaluations on SlotTransport are done with five objects using images from a top-view camera. Experiments involving single-step dynamics, long-horizon predictions, and object rearrangements are on scenes with three objects with an angled camera.

\textit{Real-world:} We use a UR5e robot with a custom-printed cylindrical end-effector and an RGB camera. We collect data $\sim 20$ episodes $\times 40$ steps of random pushes for subsets of 3 real YCB objects. Models trained in the simulation for the same object subset are retrained on this real-world data.

\subsection{Baselines}
\noindent We evaluate our approach against various methods:

\noindent\textit{Object Discovery:} Our SlotTransport, is compared with the original slot attention approach \cite{locatello2020object}. We follow the implementation of this baseline by excluding the transport mechanism introduced in our SlotTransport during training. Furthermore, we compare our approach with the off-the-shelf SAM \cite{kirillov2023segment}.

\noindent\textit{Multi-Object Dynamics:} For action-conditioned graph-based dynamics, we consider ForwGNN \cite{ye2020object}, which uses supervision of ground-truth object masks. The scene graph's nodes are embedded with the ground-truth object positions and masks to directly reconstruct the future image. We also compare with KINet \cite{rezazadeh2023kinet}, an unsupervised model that determines dynamics by identifying a set of keypoints from the scene image. Lastly, we compare with the SlotMLP variant. While it utilizes slots from SlotTransport, it models dynamics with MLPs rather than the graph-based approach of SlotGNN.

\noindent\textit{Evaluation Metrics:} We compute pixel-wise mean squared error (MSE) and Learned Perceptual Image Patch Similarity (LPIPS) \cite{zhang2018unreasonable} to measure the accuracy of the slots in reconstructing the scene composition. Additionally, to quantify the quality and consistency of the slot masks, we compute the mean Intersection over Union (IoU) for slot masks produced by SlotTransport, SlotAttention, and SAM, comparing them against the ground-truth masks from simulation.


\begin{figure}[t]
    \centering
    \includegraphics[width=\columnwidth]{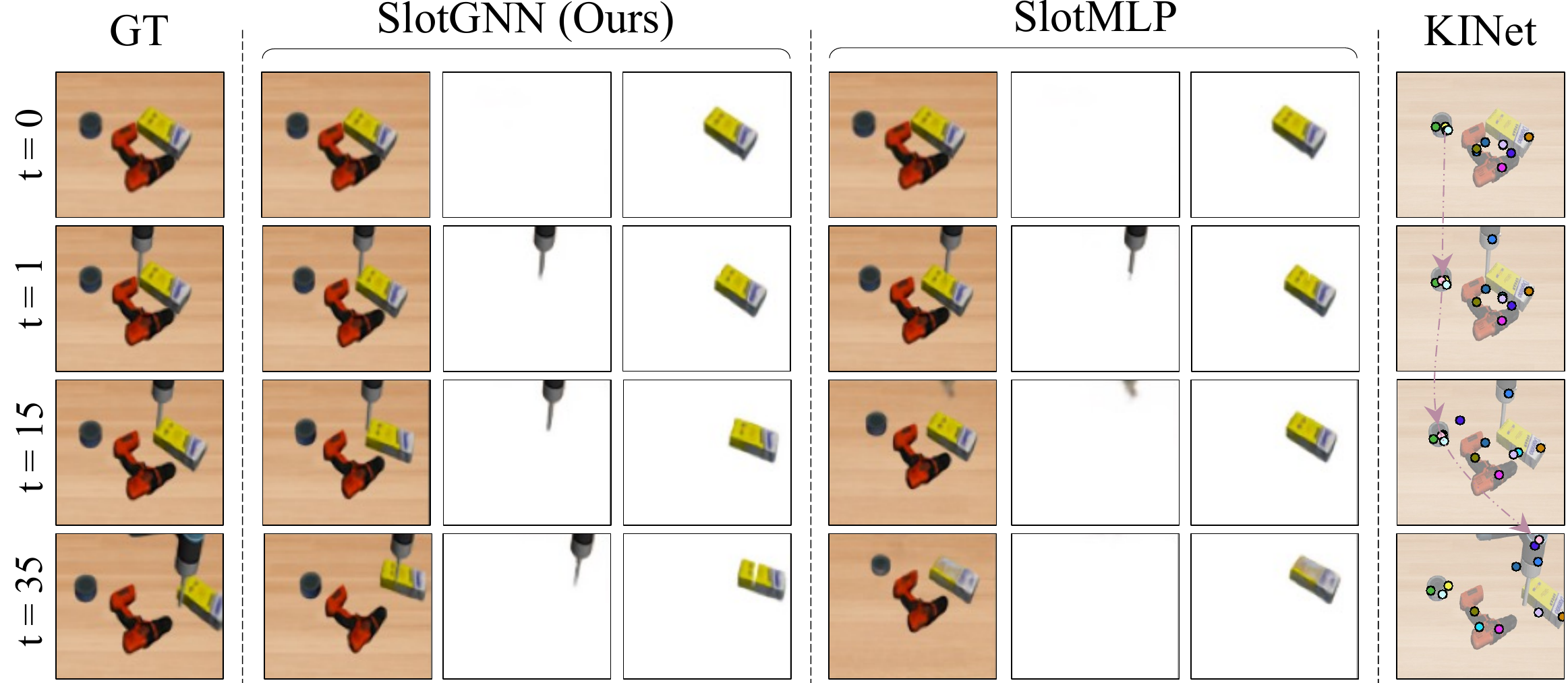}
    \caption{Long-horizon slot dynamics prediction: SlotGNN has more stability compared to SlotMLP. We also show keypoints detected with KINet \cite{rezazadeh2023kinet}.}
    \label{fig:long-horizon-qual}
 \vspace{1.2em}
    \centering
    \includegraphics[width=0.9\columnwidth]{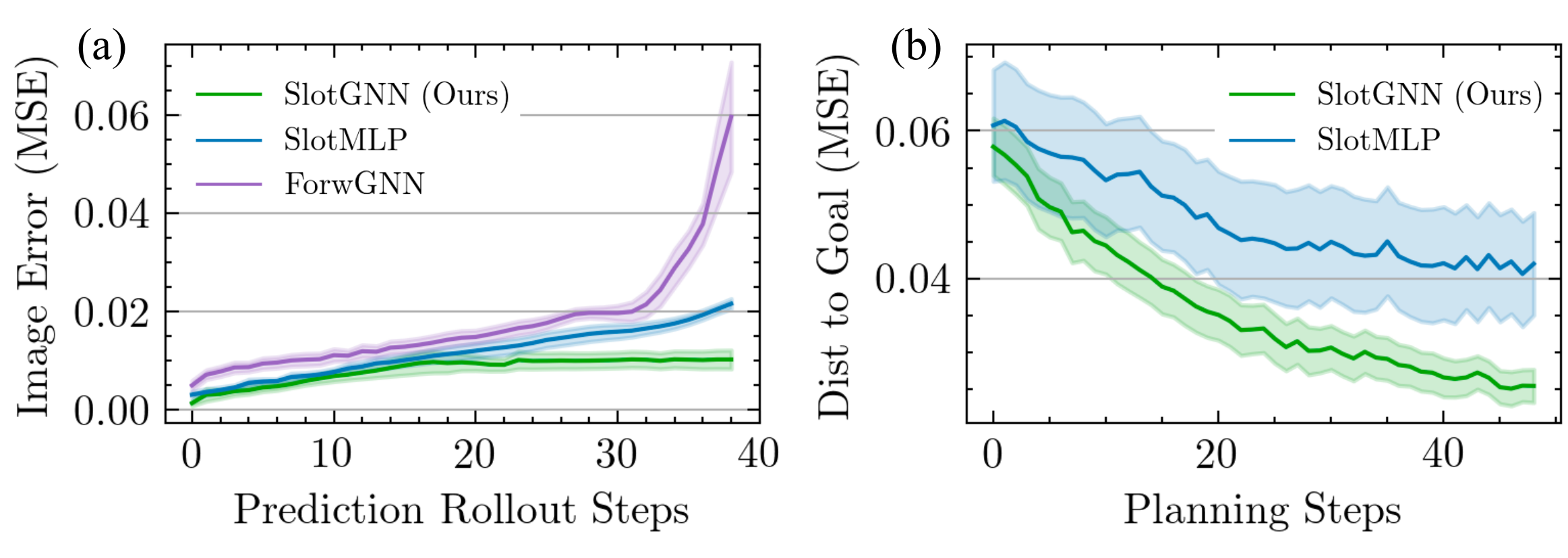}
    \vspace{-2mm}
    \caption{(a) Long-horizon dynamics rollout error. SlotGNN exhibits robust dynamics predictions as the timestep increases. (b) Planning results: Comparing the distance to the goal image between SlotGNN and SlotMLP.}
\label{fig:longhorizon-mpc}
 \vspace{1.2em}
    \centering
    \includegraphics[width=1\linewidth]{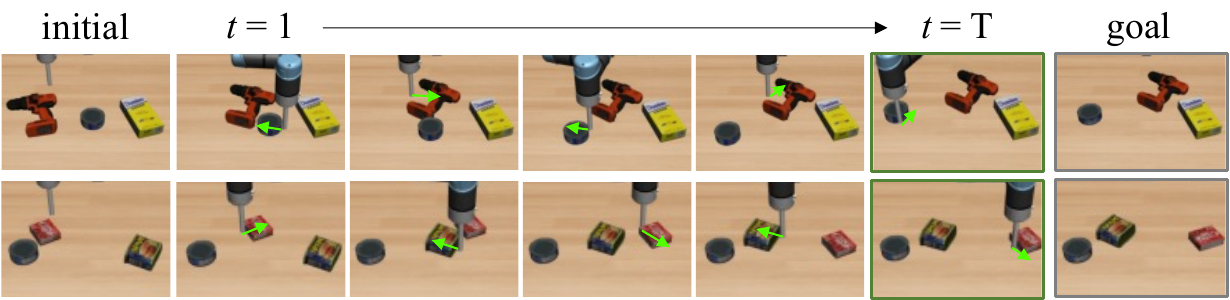}
    \caption{Qualitative results on control. Each row shows the action sequence (highlighted in green) optimized to maximize scene similarity to goal image.}
    \label{fig:plan-qual}
    \vspace{-2mm}
\end{figure}

\section{Results}
\subsection{Object Discovery Performance}

Figure \ref{fig:segment-qual} showcases the slot masks and slot reconstructions. In a scene with five objects, SlotTransport qualitatively outperforms the SlotAttention baseline \cite{locatello2020object}. SlotTransport accurately identifies all distinct visual elements, and predicts an accurate mask for each—even under heavy occlusion. However, as seen in Fig. \ref{fig:segment-qual}, the SlotAttention baseline overlooks the spam object occluded by the power drill. Moreover, SlotTransport delineates clear boundaries for each slot and accurately reconstructs their appearance. In contrast, the SlotAttention baseline presents indistinct, blurred object masks and reconstructions. We further show that relying on off-the-shelf segmentation methods, such as SAM \cite{kirillov2023segment}, is not optimal for learning object representations. This is primarily due to SAM's tendency to over-segment textured objects (e.g., backgrounds) and under-segment cluttered objects. 

Table \ref{tab:segment-quant} summarizes the quantitative evaluation of both the visual quality of reconstructed slots and the precision of slot masks. SlotTransport distinctly outperforms the SlotAttention baseline by achieving significantly better visual fidelity, measured in MSE and LPIPS. Furthermore, object masks produced by SlotTransport demonstrate superior alignment with ground-truth masks derived from simulated data. In contrast, SlotAttention often struggles to align slots accurately to cluttered objects, as shown in Fig \ref{fig:segment-qual}. This limitation is evident in the lower mIoU for SlotAttention compared to SlotTransport.

\begin{table}[t]
    \centering
    \caption{
    \begin{flushleft}
    \vspace{-2mm}
    Object discovery performance measured as visual quality (MSE and LPIPS) and mask consistency (mIoU) (MSE and LPIPS values are scaled $\times10^{-2}$).
    \vspace{-2mm}
    \end{flushleft}
    }
    \begin{tabular}{lccc}
    \toprule
    \textbf{Method} & \textbf{MSE} $\downarrow$ & \textbf{LPIPS} $\downarrow$ & \textbf{mIoU} (\%) $\uparrow$ \\
    \midrule
    \textbf{SlotTransport (Ours)}  & \textbf{0.17 ± 0.06}        & \textbf{1.29 ± 0.46}            & \textbf{93.4 ± 0.7}\\
    SlotAttention \cite{locatello2020object}
                            & 0.43 ± 0.17         & 4.79 ± 0.32             & 50.5 ± 16.2\\
    SAM  \cite{kirillov2023segment}
                            & N.A.         & N.A.            & 86.9 ± 8.5\\
    \bottomrule
    \end{tabular}
    \label{tab:segment-quant}
\end{table}

\begin{table}[t]
    \centering
    \caption{
        \begin{flushleft}
\vspace{-2mm}
Single-step dynamics prediction accuracy measured as visual quality (MSE) and mask consistency (mIoU).
\vspace{-2mm}
    \end{flushleft}
    }
    \begin{tabular}{lccc}
    \toprule
    \textbf{Method} & \textbf{Supervision} & \textbf{MSE} $\downarrow$ & \textbf{mIoU} (\%) $\uparrow$ \\
    \midrule
    \textbf{SlotGNN (Ours)}  & \textbf{Img }      & \textbf{0.14 ± 0.05} & \textbf{86.9 ± 2.9} \\
    SlotMLP         & Img       & 0.32 ± 0.09 & 72.6 ± 1.1 \\
    KINet  \cite{rezazadeh2023kinet}
                    & Img       & 1.86 ± 0.09 & N.A.        \\
    ForwGNN \cite{ye2020object}        
                    & GT State       & 0.50 ± 0.14 & N.A. \\
    \bottomrule
    \end{tabular}
    \label{tab:onestep-dyn}
    
\end{table}

\begin{figure*}[t]
    \centering
    \includegraphics[width=0.9\linewidth]{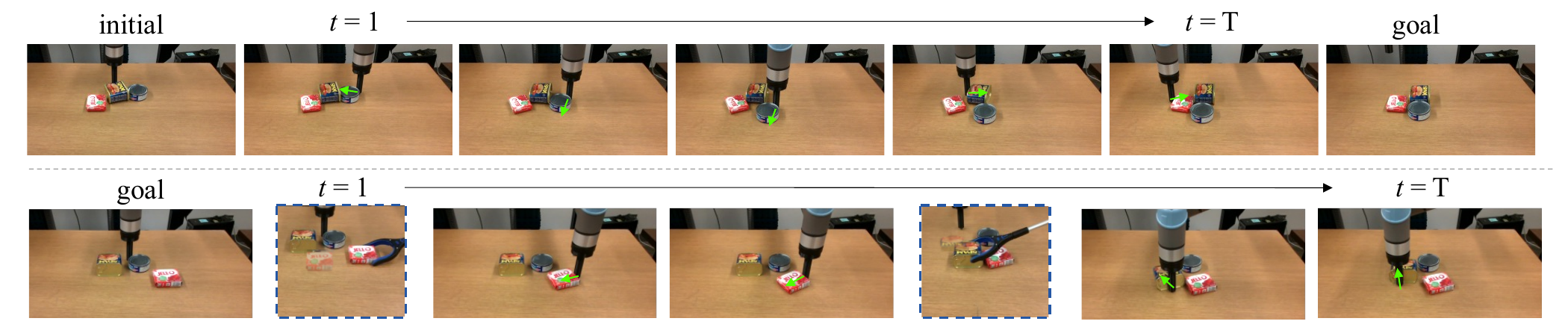}
    \caption{Real-world control using SlotTransport and SlotGNN: The top row shows objects being rearranged to align with a goal image. In the bottom row, objects are persistently displaced from their goal positions, the robot comes up with a sequence of actions to push the objects back to their desired locations. Please visit our project page for videos and more examples: \href{http://bit.ly/slotgnn}{bit.ly/slotgnn}.}
    \label{fig:plan-qual-real}
\end{figure*}

\begin{figure}[t]
    \centering
    \includegraphics[width=1\columnwidth]{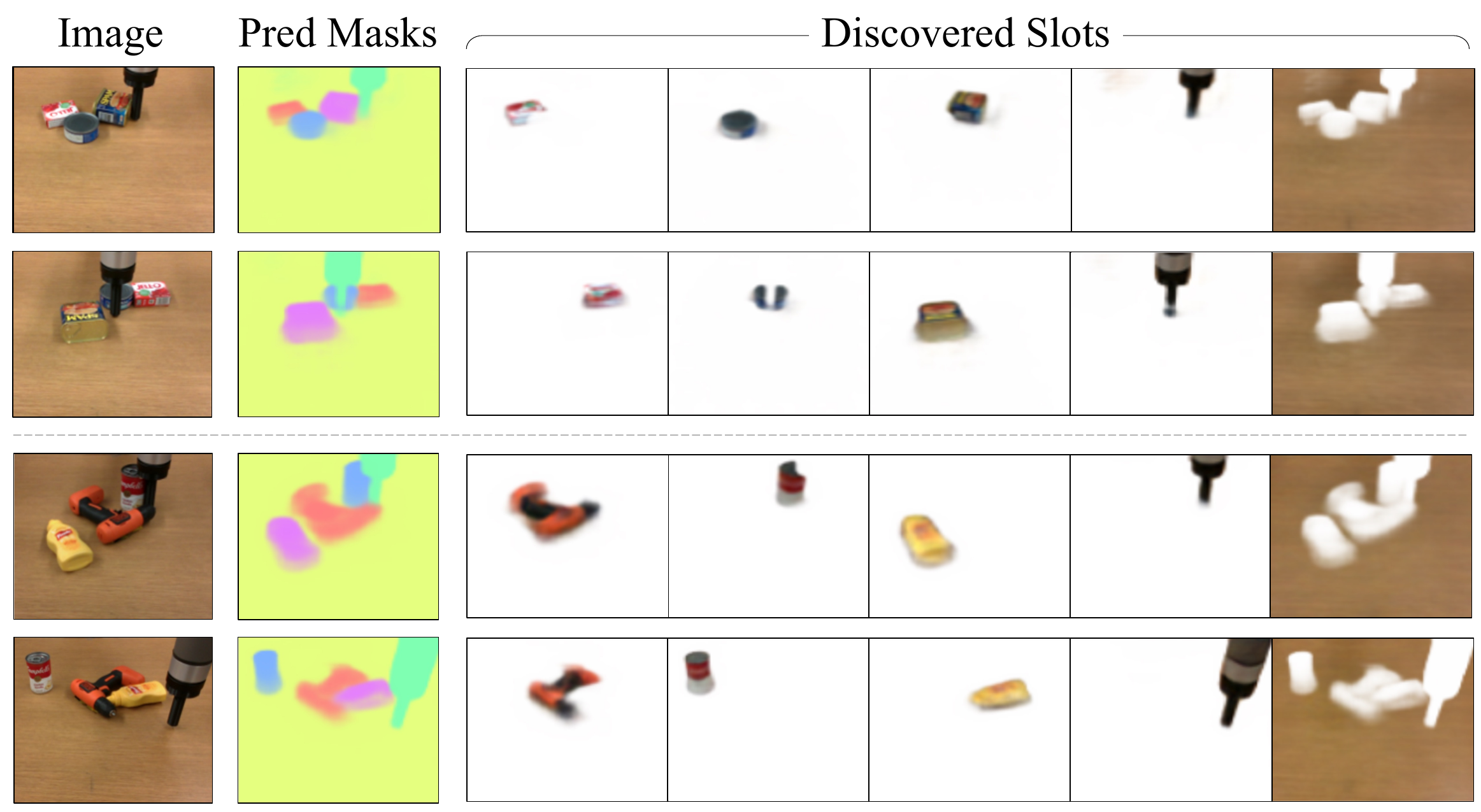}
    \caption{
    Real-world object slot discovery with SlotTransport. Our unsupervised framework transfers to real settings and discovers accurate object-centric representations that reflect the positional and visual features of the objects.
    }
    \label{fig:real-slots}
\end{figure}

\subsection{Dynamics Prediction Performance}

Figure \ref{fig:qual-dyn} illustrates the single-step dynamics prediction of SlotGNN. By taking as input the current image and the intended robot's pushing action vector, our model accurately predicts the future scene. It does so by predicting the future state of each slot, based on the learned multi-object dynamics of the scene. The quantitative results presented in Table \ref{tab:onestep-dyn} highlight the accuracy of SlotGNN in single-step dynamics prediction. In single-step dynamics prediction, SlotGNN outperforms all other baselines, including the SlotMLP variant and the unsupervised keypoint dynamics KINet \cite{rezazadeh2023kinet}. It's worth noting that while ForwGNN does rely on ground-truth state information for supervision, it still falls short in MSE compared to SlotGNN, which utilizes image-based supervision. This further highlights the robustness of the detected slots in SlotTransport in representing objects enabling SlotGNN to learn accurate multi-object dynamics.

As illustrated in Fig. \ref{fig:long-horizon-qual}, SlotGNN excels in predicting stable long-horizon dynamics compared with SlotMLP. Although the scenes reconstructed with SlotGNN may diverge from the ground-truth due to cumulative prediction errors, it yields physically plausible future scenes. In contrast, SlotMLP struggles to retain the coherence slots over time. Given that both SlotGNN and SlotMLP use slots by SlotTransport, the difference in their long-horizon predictions can be attributed to the graph-based model's enhanced ability to capture multi-object dynamics. In Fig. \ref{fig:long-horizon-qual}, unsupervised keypoints detected by KINet \cite{rezazadeh2023kinet} are also shown. KINet requires stable keypoint-object correspondences to learn multi-object dynamics. This stability is compromised when a robot enters or exits the frame or introduces object occlusions (see the pink keypoint in the last column of Fig. \ref{fig:long-horizon-qual}). A quantitative summary of the long-horizon rollout outcomes can be found in Fig. \ref{fig:longhorizon-mpc}-a. 


\subsection{Planning with SlotGNN}

Fig. \ref{fig:plan-qual} shows our method's application in control tasks. In a challenging object rearrangement scenario, the robot plans an action sequence using SlotTransport and SlotGNN. Through accurate multi-object dynamics projections, the robot effectively aligns objects to a desired configuration using just the RGB image. The planning performance of slot-based models are compared in Fig. \ref{fig:longhorizon-mpc} which emphasizes the effectiveness of a graph-based model in learning object-centric dynamics.

\subsection{Real-World Experiments}

Demonstrating the real-world applicability of our unsupervised approach, we successfully transfer SlotTransport and SlotGNN, initially trained in simulation, to the real robot by collecting a minimal dataset of just 20 real robot demonstrations (5\% of the amount of simulated training data). SlotTransport retains its accuracy in the real environment as shown in Fig \ref{fig:real-slots}. The slots discovered from the real mutli-object scene, clearly distinguish all the scene elements even under occlusion. For the real-world control, we experiment with two tasks as shown in Fig. \ref{fig:plan-qual-real}. The first scenario, presented in the top row, involves rearranging objects to achieve a predetermined goal image. The bottom row showcases a more dynamic scenario where objects are continuously displaced from their target positions by a human with a grabber stick. In response, our robot, using SlotTransport and SlotGNN, finds a sequence of actions to restore the objects to their intended locations.

\section{Conclusion}
This work addresses the challenges of unsupervised learning for multi-object dynamics through visual observations. We present SlotTransport, a novel approach based on slot attention for unsupervised object discovery, ensuring temporal consistency in object-centric representations. Alongside this, we introduce SlotGNN, an unsupervised graph-based dynamics model for predicting the future states of multi-object scenes using the slots. Both methods have proven effective in complex robotic control tasks and long-horizon dynamics prediction. Importantly, we demonstrate that our unsupervised approach, using SlotTransport and SlotGNN, successfully transfers to real-world settings and enables object discovery and dynamic modeling solely from RGB images. For limitations, one key aspect we recognize is that our slot discovery process currently necessitates pre-determining the number of slots. In our experiments, we predefined the slot count equal to the anticipated number of elements in the scene. Developing a more adaptive mechanism that automatically determines the required slot count could be a promising future research direction.

\section{Acknowledgement}
We thank Carl Winge for the help with the robot setup, Chahyon Ku for providing helpful feedback on our initial draft, and all other members of the Robotics Perception and Manipulation Lab for their insightful discussions. This project is partially funded by the UROP Program at the University of Minnesota and the MnDRIVE UMII (University of Minnesota Informatics Institute) Seed Award. This project was also supported in part by the Sony Research Award Program and NSF Award 2143730

\bibliographystyle{IEEEtran} 
\bibliography{references}


\end{document}